\documentclass[10pt,twocolumn,letterpaper]{article}

\usepackage{iccv}
\usepackage{times}
\usepackage{epsfig}
\usepackage{graphicx}
\usepackage{amsmath}
\usepackage{amssymb}

\usepackage{algorithm}
\usepackage{algorithmicx}
\usepackage{algpseudocode}

\usepackage{booktabs}
\usepackage{tabularx}
\newcolumntype{Y}{>{\centering\arraybackslash}X}
\usepackage{xspace}
\usepackage{amsmath}
\usepackage{enumitem}
\usepackage[breaklinks=true,bookmarks=false,pagebackref=true]{hyperref}

\iccvfinalcopy % *** Uncomment this line for the final submission

\def\iccvPaperID{582} % *** Enter the ICCV Paper ID here

\ificcvfinal\fi

\begin{document}

%%%%%%%%% TITLE
\title{ShellNet: Efficient Point Cloud Convolutional Neural Networks using \\Concentric Shells Statistics}
%\title{ShellNet: Efficient Point Cloud CNNs using Concentric Shells Statistics}

\author{
Zhiyuan Zhang$^{1}$ \hspace{0.2in}
Binh-Son Hua$^{2}$ \hspace{0.2in}
Sai-Kit Yeung$^{3}$
\vspace{0.1in}\\
$^{1}$Singapore University of Technology and Design \quad
$^{2}$The University of Tokyo
\\
$^{3}$Hong Kong University of Science and Technology
%\\
%{\tt\small cszyzhang@gmail.com \tt \tt \quad \small binhson.hua@gmail.com \tt \tt \quad \small saikit@ust.hk}
}

\maketitle
% Remove page # from the first page of camera-ready.
\ificcvfinal\thispagestyle{empty}\fi

\def\ournet{ShellNet\xspace}
\def\ourconv{ShellConv\xspace}
\def\smallgap{\vspace{0.05in}}
\newcommand{\new}[1]{#1}
%%%%%%%%% ABSTRACT
\begin{abstract}
   Deep learning with 3D data has progressed significantly since the introduction of convolutional neural networks that can handle point order ambiguity in point cloud data. While being able to achieve good accuracies in various scene understanding tasks, previous methods often have low training speed and complex network architecture. In this paper, we address these problems by proposing an efficient end-to-end permutation invariant convolution for point cloud deep learning. Our simple yet effective convolution operator named \ourconv uses statistics from concentric spherical shells to define representative features and resolve the point order ambiguity, allowing traditional convolution to perform on such features. Based on \ourconv we further build an efficient neural network named \ournet to directly consume the point clouds with larger receptive fields while maintaining less layers. We demonstrate the efficacy of \ournet by producing state-of-the-art results on object classification, object part segmentation, and semantic scene segmentation while keeping the network very fast to train. Our code is publicly available in our project page \footnote{\url{https://hkust-vgd.github.io/shellnet/}}.
\end{abstract}

%%%%%%%%% BODY TEXT
\section{Introduction}
\label{indroduction}

Convolutional neural networks (CNNs) have shown significant success in image and pattern recognition, video analysis, and natural language processing~\cite{lecun2015deep}. %However, most previous works focus on 2D signals. Recently, 3D deep learning 
Extending this success from 2D to 3D domain has been receiving great interests. Promising results have been demonstrated for the long-standing problem of scene understanding. Previously a 3D scene is often represented using structured representations such as volumes \cite{qi2016volumetric,li2016fpnn}, multiple images~\cite{su2015multi,qi2016volumetric}, hierarchical data structures \cite{riegler2017octnet,klokov2017escape,wang2017cnn}. However, such representations usually face great challenges from memory consumption, imprecise representation, or lack of scalability for tasks such as classification and segmentation.  

\begin{figure}[t]
	\centering
	\includegraphics[width=0.49\linewidth]{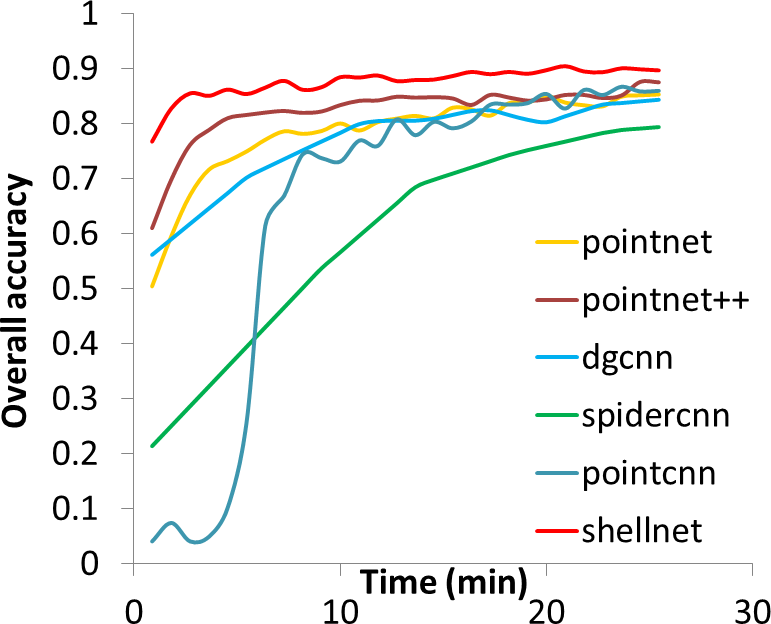}
	\includegraphics[width=0.49\linewidth]{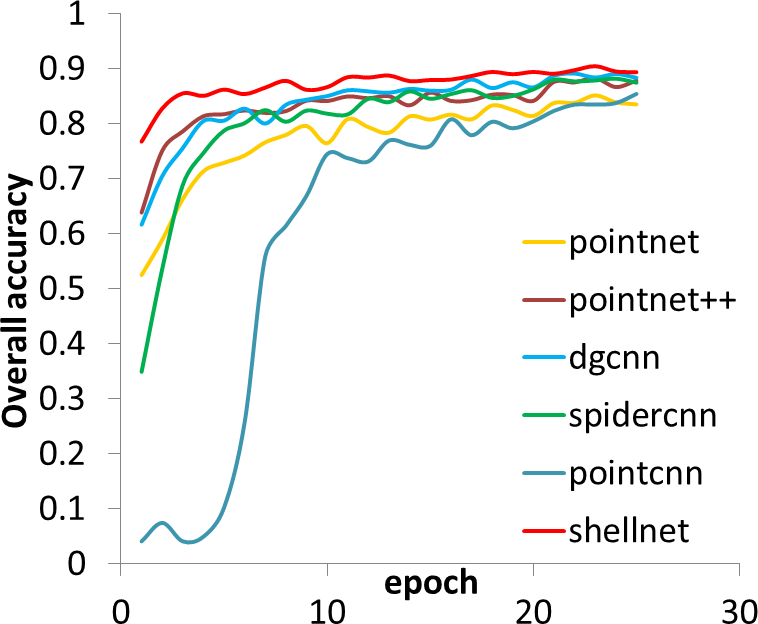}
	\label{fig:time}
	\caption{The accuracy of point cloud classification of different methods over time and epochs. While being accurate, some methods are quite costly to train. We address this problem by \ourconv, a simple yet effective convolutional operator based on concentric shell statistics. In both equal-time and equal-epoch comparisons, our method performs the best. It can achieve over 80\% accuracy within two minutes, and reach 90\% on the test dataset after only 15 minutes of training.} 
\end{figure}

Recently, directly consuming point clouds using neural networks has shown great promises ~\cite{qi2017pointnet,qi2017pointnet++,xu2018spidercnn,li2018pointcnn}. PointNet~\cite{qi2017pointnet} pioneers this direction by learning with a symmetric function to make the network robust to point order ambiguity. Many subsequent works extend this direction by designing convolution that better captures local features of a point cloud. While such efforts lead to improved scene understanding performance, there is often a trade-off between network complexity, training speed, and accuracy. For example, the follow-up work PointNet++ \cite{qi2017pointnet++} segments point cloud into smaller clusters and applies PointNet locally in a hierarchical manner. While achieving better result, the network is more complicated with reduced speed. Pointwise convolution~\cite{hua2017point} is simple to implement but inaccurate. SpiderCNN~\cite{xu2018spidercnn} extends traditional convolution on 2D images to 3D point clouds by parameterizing a family of convolution filters. Although high accuracy is achieved, more time is taken for training. PointCNN~\cite{li2018pointcnn} achieves the state-of-the-art accuracy via learning a local convolution order but its training is slow to converge. In general, designing a convolution for point cloud that can strike a good balance between such performance factors is a challenging problem.
%The extracted features globally describe the input point cloud that can be applied for various computer vision tasks. To better capture local features, the follow-up work PointNet++ \cite{qi2017pointnet++} segments point cloud into smaller clusters and applies PointNet onto these clusters in a hierarchical manner. While achieving better result, the network structure becomes more complicated with reduced speed. SpiderCNN~\cite{xu2018spidercnn} extends traditional convolution on 2D images to 3D point clouds by parameterizing a family of convolution filters. Although high accuracy is achieved on classification and part segmentation, more time is taken for training. Also, both PointNet++ ~\cite{qi2017pointnet++} and SpiderCNN ~\cite{xu2018spidercnn} incorporate normals to achieve their best results. A more recent work PointCNN ~\cite{li2018pointcnn} presents an $\chi$ operator to extract local features from the point cloud. It uses multilayer perceptron to learn the local point order based on which the traditional convolution can be performed. A convolutional network is then constructed with $ \chi$ operator as core with high accuracies shown using pure 3D geometric points as input. Despite the fast running every epoch, the convergence speed is slow down making the training inefficient. 

Based on these observations, we propose a novel approach to consume point clouds directly in a very simple neural network which is able to achieve the state-of-the-art accuracy with very fast training speed, as shown in Figure~\ref{fig:time}. 
Our idea is to split a local point neighborhood such that point neighboring and convolution with points can be performed efficiently.
To achieve this, at each point, we query the point neighborhood and partition it with a set of concentric spheres, resulting in concentric spherical shells. In each shell, the representative features can be extracted based on the statistics of the points inside. By using \ourconv as the core convolution operator, an efficient neural network called \ournet can be constructed to solve 3D scene understanding tasks such as object classification, object part segmentation, and semantic scene segmentation.

In general, the main contributions of this work are:
\begin{itemize}[leftmargin=*]
%\item Shell structures are defined locally to divide the local patch into concentric shells where the convolution order can be naturally defined.

\item \ourconv, a simple yet effective convolution operator for orderless point cloud. The convolution is defined on a domain that can be partitioned by concentric spherical shells, simultaneously allowing efficient neighbor point query and resolving point order ambiguity by defining a convolution order from the inner to the outer shells;

\item \ournet, an efficient neural network architecture based on \ourconv for learning with 3D point clouds directly without having any point order ambiguity;

\item Applications of \ournet on object classification, object part segmentation, and semantic scene segmentation that achieves the state-of-the-art accuracy.
\end{itemize}

%To facilitate reproducibility, we also provide a thorough discussion about the implementation and the quantitative experiments. Our network can extract features effectively that reflect the , and thus is very fast to train. 

%-------------------------------------------------------------------------
\section{Related Works}
\label{related_works}
Recent advances in computer vision witness the growing availability of 3D scene datasets~\cite{armeni-parsing-cvpr16, wu-3dshapenets-cvpr15, yi2016scalable}, leading to deep learning techniques to tackle the long-standing problem of scene understanding, particularly object classification, object part and scene segmentation. 
In this section, we review the state-of-the-art research in deep learning with 3D data, and then focus on techniques that enable feature learning on point clouds for scene understanding tasks. 

%\paragraph{Learning with Volumes and Multi-View Images.}
Early deep learning with 3D data uses regular representations such as volumes \cite{wu20153d,maturana-voxnet-iros15,qi2016volumetric,li2016fpnn} and multi-view images \cite{su2015multi,qi2016volumetric} for feature learning to solve object classification and semantic segmentation. 
Unfortunately, volume representation is very limited due to large memory footprints. 
Multi-view image representation does not have this issue, but it stores depth information implicitly, which makes it challenging to learn view-independent features. 

\new{
Recently, deep learning in 3D focuses toward point clouds, which is more compact and intuitive compared to volumes. As point cloud is mathematically a set, using point cloud with deep neural networks requires fundamental changes to the core operator: convolution. Defining efficient convolution for point clouds has since been a challenging, but an important task. 
Inspired from learning with volumes,
Hua et al.~\cite{hua2017point} perform on-the-fly voxelization at each point of the point cloud based on nearest point queries. Le et al.~\cite{le2018pointgrid} propose to apply convolution on a regular grid with each cell containing point features that are resampled to a fixed size. Tatarchenko et al. ~\cite{tatarchenko2018} perform convolution on the local tangent planes. 
Xie et al.~\cite{xie2018shapecontext} generalize shape context to convolution for point cloud. Liu et al.~\cite{liu2018point2seq} use a sequence model to summarize local features with multiple scales. Such techniques lead to straightforward implementations of convolutional neural network for point clouds. However, extra computations are required for the explicit data representation, making the learning inefficient.
}

\new{
%\paragraph{Learning with Point Clouds.}
Instead of voxelization, it is possible to make neural network operate directly on point clouds. Qi et al.~\cite{qi2017pointnet} propose PointNet, a pioneering network that learns global per-point features by optimizing a symmetric function to achieve point order invariance. The drawback of PointNet is that each point feature is learnt globally, i.e., no features from local regions are considered. 
Recent methods in point cloud learning are focused on designing convolution operators that can capture such local features.
}

\new{
In this trend, PointNet++~\cite{qi2017pointnet++} supports local features by a hierarchy of PointNet, and relies on a heuristic point grouping to build the hierarchy. 
Li et al.~\cite{li2018pointcnn} propose to learn a transformation matrix to turn the point cloud to a latent canonical representation, which can be further processed with standard convolutions.
Xu et al.~\cite{xu2018spidercnn} propose to parameterize convolution kernels with a step function and Taylor polynomials. 
Wang et al.~\cite{wang2018edgeconv} propose a similar network structure to PointNet by optimizing weights between a point and its neighbors and using them for convolution. 
Shen et al.~\cite{shen2018mining} also improve a PointNet-like network by kernel correlation and graph pooling. 
Huang et al.~\cite{huang2018recurrent} learn the local structure particularly for semantic segmentation by applying traditional learning algorithms from recurrent neural networks. 
%Despite the improvements, the receptive fields of convolution are still restricted to the individual points. 
Ben-Shabat et al.~\cite{shabat20183dmfv} use a grid of spherical Gaussians with Fisher vectors to describe points.
Such great efforts lead to networks with very high accuracies, but the efficiency of the learning is often overlooked (see Figure~\ref{fig:time}). This motivates us to focus on efficiency for local features learning in this work.
} 

\begin{figure*}[t]
	\centering
	\includegraphics[width=0.8\linewidth]{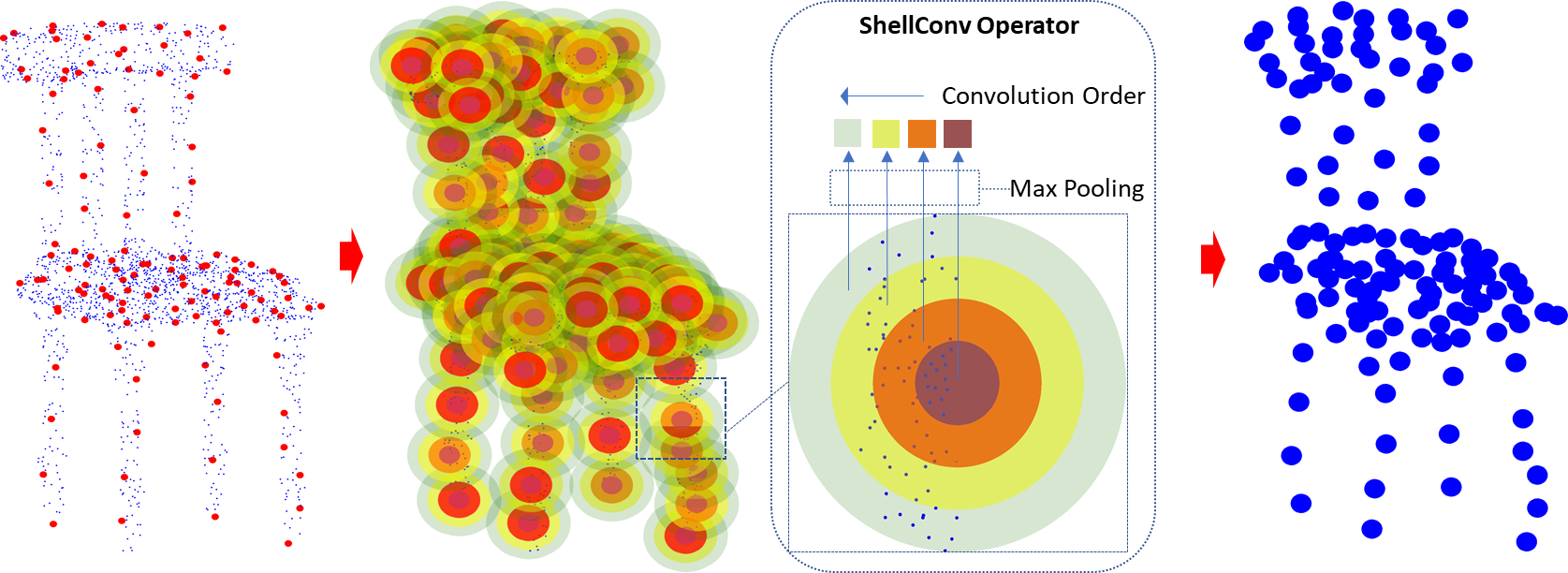}
	\parbox[b]{1\textwidth}{\relax
		\quad \quad \quad \quad \qquad \qquad (a) \qquad \qquad \quad \qquad \qquad  (b) \quad \qquad \qquad \qquad \qquad (c)\qquad \qquad \quad \quad \quad\quad\quad\quad\quad\quad(d)}
	%	\caption{\ourconv operator applied at a point $p$. (a) $K$ neighbors of point $p$ are queried to yield a feature vector of $K \times f_0$, where $f_0$ is number of input data channels. (b) The point set is transformed into a local coordinate frame at $p$. (c) Point features (illustrated by thicker points) for use in convolution is established by two parts: the input feature vector $K \times f_0$, and the point features of size $K \times f_1$ obtained by lifting point coordinates $K \times 3$ lifted to a high-dimensional space by a shared multi-layer perceptron. This ensures that the convolution takes into account point coordinates and both input point features. (d) Concentric spherical shells at $p$ are now considered. The point distribution in each shell is represented by a statistical approach. Here we choose to only use the maximum value of the distribution. The maxpooled features $m_0$, $m_1$, $m_2$, $m_3$ are concatenated with features at point $p$, resulting in a fixed size feature vector. This vector is transformed by a standard convolution to yield the output of the \ourconv operator.}
	\caption{ShellConv operator. (a) For an input point cloud with/without associated features, representative points (red dots) are randomly sampled. The nearest neighbors are then chosen to form a point set centered at the representative points. 
		%and each is centralised with the associated representative point to transform into local coordinate frame. The points covered under the local patch are lifted into a high dimensional space by a shared multi-layer perceptron. 
		The point sets are distributed across a series of concentric spherical shells (b) and the statistics of each shell is summarized by a maxpooling over all points in the shell, the features of which are lifted by an \textit{mlp} to a higher dimension. The maxpooled features are indicated as squares with different colors (c). Following the inner to the outer order, a standard 1D convolution can be performed to yield the output features (d). Thicker dot means less points but each has higher dimensional features. 
		%Note that, the radius of sphere is arithmetic progress. In practice, the shells would not so equidistant as is shown here when the data is non-uniformly distributed. 
	}
	\label{fig:conv}
\end{figure*}

\new{
Beyond learning on unstructured point clouds, there have been some notable extension works, such as learning with hierarchical structures \cite{riegler2017octnet,klokov2017escape,wang2017cnn,wang2018aocnn},  
%Such techniques could be limited due to their slow inference, quantization artifacts, and information lost.
learning with self-organizing network~\cite{li2018sonet},
learning to map a 3D point cloud to a 2D grid~\cite{yang2018foldingnet, groueix2018atlasnet}, addressing large-scale point cloud segmentation~\cite{landrieu-superpoint-cvpr18}, handling non-uniform point cloud~\cite{hermosilla2018monte}, and employing spectral analysis~\cite{yi2017syncspeccnn}.  
Such ideas are orthogonal to our method, and adding them on top of our proposed convolution could be an interesting future research. 
}

%-------------------------------------------------------------------------
\section{The \ourconv Operator}
\label{loc_cov}

To achieve an efficient neural network for point cloud, the first task is to define a convolution that is able to directly consume a point cloud. Our problem statement is given a set of points as input, define a convolution that can \emph{efficiently} output a feature vector to describe the input point set. 

There are two main issues when defining this convolution. First, the input point set has to be defined. It can be the entire point cloud, or a subset of the point cloud. The former case seeks a global feature vector that describes the entire point cloud; the latter seeks a local feature vector for each point set that can be further combined when needed.
Second, one has to seamlessly take care of the point order ambiguity in a set and the density of the points in the point cloud. 
PointNet~\cite{qi2017pointnet} opted to learn global features, but it has been shown by recent works~\cite{qi2017pointnet++,li2018pointcnn,wang2018edgeconv,xu2018spidercnn} that local features can lead to more representative features, resulting in better performance. We are motivated by these works and define a convolution to obtain features for a local point set. To keep our convolution simple but efficient, we propose an intuitive approach to addresses the challenges, below.

\begin{algorithm}[t]
	\caption{ \ourconv Operator.}
	\label{alg:conv}
	\begin{algorithmic}[1]
		\Require
		\\$p$, $\Omega_p$, $\{ F_{prev}(q) : q \in \Omega_p \}$ \hfill * Representative point, point set, and previous layer features of point set.
		\Ensure
		$F_{p}$ \hfill * Convolutional features of $p$.
		\State $\{ q \} \leftarrow \{ q - p : \forall q \in \Omega_p \}$  \hfill * Neighbor point $q$ is localised with $p$ as the center.
		\State $\{ F_{local}(q) \} \leftarrow \{ \textrm{mlp}(q) \}$ \hfill * Individually lift each point $q$ to a higher dimensional space.
		\State $\{ F(q) \} \leftarrow \{ [F_{prev}(q), F_{local}(q) \}$ \hfill * Concatenate the local and previous layer features.
		\State $\{ S \} \leftarrow \{ S : q \in \Omega_S \}$ \hfill * Determine which shell $q$ belongs to according the distances from $q$ to center $p$.
		\State $\{ F(S) \} \leftarrow \{ \mathrm{maxpool}(\{ F(q) : q \in \Omega_S \})  : \forall S \}$ \hfill * Get fixed-size feature of each shell by a maxpool over all points in the shell.
		\State $F_{p} \leftarrow \textrm{conv}(\{F(S)\})$ \hfill * Perform a 1D convolution with all shell features from inner to outer.
		\\
		\Return $F_{p}$
	\end{algorithmic}
\end{algorithm}
%\begin{equation}
%\label{eq-1}
%F_{c} = ShellConv(p, P, F_{prev}) = Conv()
%\end{equation}

\begin{figure*}[t]
	\centering
	\includegraphics[width=0.8\linewidth]{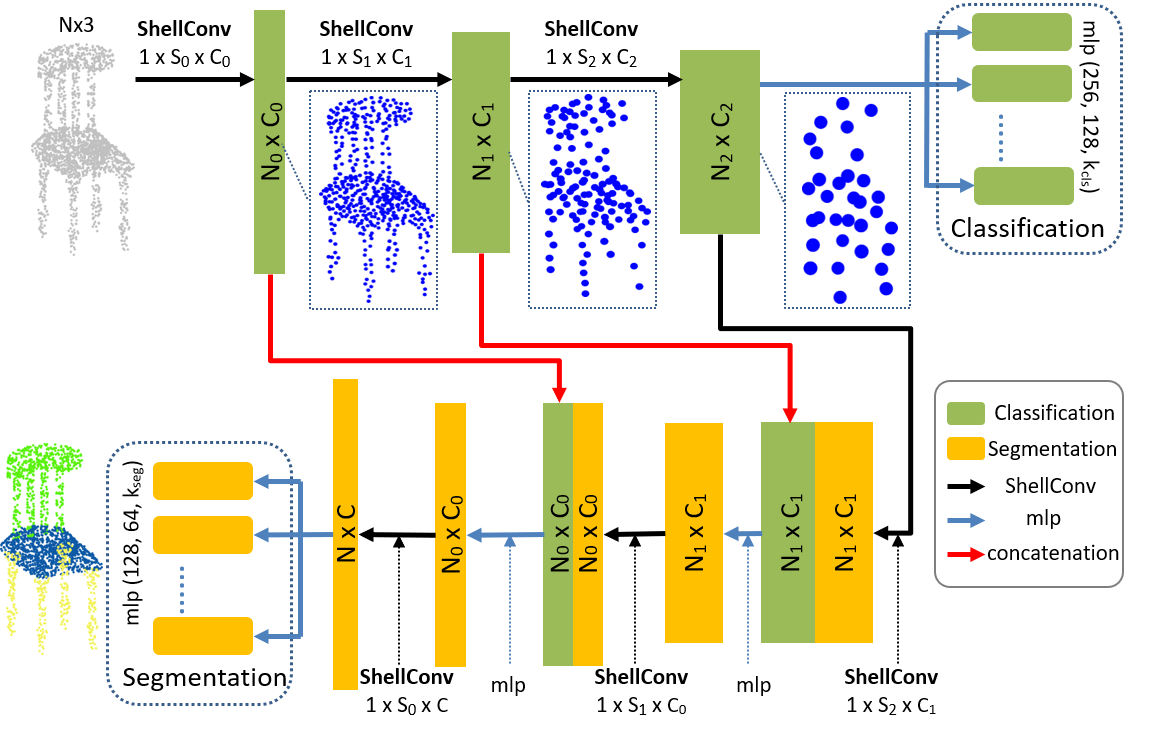}	
	\caption{\ournet architecture. For classification, we apply three layers of \ourconv before the fully connected classifier. For semantic segmentation, we follow a U-net~\cite{ron2015unet} architecture. The encoder is in green and the decoder is in yellow. Point downsampling and upsampling is also included in our convolution, depending on its use. $N_0 > N_1 > N_2$ denotes the number of points in the input and after being sampled in each convolution, and $C < C_0 < C_1 < C_2$ denotes the output feature channels at each point. $S_0 > S_1 > S_2$ denotes the number of shells in each \ourconv operator that is analogous to the convolution kernel size. Given a fixed shell size, when the point cloud is downsampled, the number of shells also decreases.  $1 \times S_{0} \times C_{0}$ denotes a convolution that convolutes an input features using kernel $(1, S_{0})$ and output $C_0$ feature channels.}
	\label{fig:net}
\end{figure*}

\paragraph{Convolution.}
%We show the main idea of our convolution in Figure~\ref{fig:conv}. A important concept in our convolution design for neural networks is to support a large receptive field so that the context information in spatial data can be captured effectively. To achieve this, a common strategy is to decrease the spatial resolution of the input and output more feature channels at deeper layers. We support this strategy in our convolution by combining point sampling into the convolution, outputting sparser point sets at deeper layers. Particularly, from the input point set, a set of representative points are randomly sampled (red dots in Figure~\ref{fig:conv} (a)). Each representative point and its neighbor points define a point set for convolution (Figure~\ref{fig:conv} (b)). 
We show the main idea of our convolution in Figure~\ref{fig:conv}. The common strategy in a traditional CNN architecture is to decrease the spatial resolution of the input and output more feature channels at deeper layers. We also support this strategy in our convolution by combining point sampling into the convolution, outputting sparser point sets at deeper layers. Different from previous works that stack many layers to increase receptive field, our method can obtain a larger receptive field without increasing the number of layers. Particularly, from the input point set, a set of representative points are randomly sampled (red dots in Figure~\ref{fig:conv} (a)). Each representative point and its neighbor points define a point set for convolution (Figure~\ref{fig:conv} (b)). 

Let us now focus on a single representative point $p$ and its neighbors $q \in \Omega_p$, where $\Omega_p$ is the set of neighbors determined by a nearest neighbor query.
By definition, the convolution at $p$ is
\begin{align}
	F(p)^{(n)} = \sum_{q \in \Omega_p^{(n)}} w(q)^{(n)} F(q)^{(n-1)}
\end{align}
where $F$ denotes the input feature of the point set for a particular channel, $w$ is the weight of the convolution. We use superscript ${(n)}$ to denote the data or parameters of layer $n$. Note that $F(p)$ and $F(q)$ denote the features of point $p$ and $q$. They are disregarded of the order of $p$ and $q$ in the point cloud because we simply treat the point cloud as a mathematical set. 
The only issue with this convolution here is how to define the weight function. The weights have to be suitable for training, i.e., $w$ has to be discretized into a fixed-size vector of trainable parameters. Defining $w$ for each point is not practical because the points are not ordered.

To address this issue, our observation here is that we can exploit the partitioning of the neighborhood into regions such that $w$ is well defined and the output can be computed efficiently. Particularly, to facilitate neighbor queries, we use a set of \emph{multi-scale concentric spheres} to define the regions (Figure~\ref{fig:conv}(c)). The region between two spheres forms a spherical \emph{shell}. The union of the concentric spherical shells yields the domain $\Omega_p$.
Therefore, we can define our convolution as
\begin{equation}
	F(p)^{(n)} = \sum_{S \in \Omega_p^{(n)}} w_S^{(n)} F(S)^{(n-1)}
\end{equation}
Note that as the shells are naturally ordered (from the inner most to the outermost), there is no ambiguity among the shells and the convolution is well defined, with weight $w_S$ for each shell. What remains ambiguous is the order of the points in the shells. 
To address this problem, we propose a statistical approach to aggregate features of the points in each shell such that it yields an order-invariant output. Particularly, we choose to represent the features by only the maximum value in each feature channel:
\begin{equation}
	F(S) = \mathrm{maxpool}(\lbrace F(q) : q \in \Omega_S \rbrace )
\end{equation}
where $\Omega_S$ denotes a shell $S$.
Theoretically, the maximum value is a crude approximation to the underlying distribution, but because each point often has tens or hundreds of feature channels, the information from many points in the shell can still be represented.
The detailed steps of \ourconv is presented in Algorithm~\ref{alg:conv}. 

\paragraph{Spherical Shells Construction.} We use a simple heuristics approach to establish the spherical shells as follows. We first compute the distance between the neighbor points to the representative point at the center. We then sort the distances, and distribute points to shells based on their distances to the center, from inner to outer. We assign a fixed number of points to each shell, i.e., $n$ points per shell in our implementation. Particularly, we first grow a sphere from the center until $n$ points falls inside the sphere. This is the innermost shell. After that, the sphere continues to grow to collect another $n$ points that forms the second shell, and so on. 
We found that this approach of shells construction provides a good stratification of point distributions in the shells. It is also easy to implement and has low overhead. 

%\paragraph{Relevance to PointNet.}
%Our approach to convolution that uses statistics in spherical shells to resolve ambiguity in point ordering is relevant to the maxpooling operator used by PointNet~\cite{qi2017pointnet}. In their work, the maxpooling is used to aggregate a global feature vector while here, we use maxpooling to output local feature vectors. Their extension~\cite{qi2017pointnet++} stacks several PointNet to make an hierarchy to learn local features. We will show later that our much simpler approach can still learn local features effectively and achieve the state-of-the-art results.

%\paragraph{Relevance to PointCNN ~\cite{li2018pointcnn}.}
%Our approach to convolution that uses statistics in spherical shells to resolve ambiguity in point ordering is relevant to the maxpooling operator used by PointNet~\cite{qi2017pointnet}. In their work, the maxpooling is used to aggregate a global feature vector while here, we use maxpooling to output local feature vectors. Their extension~\cite{qi2017pointnet++} stacks several PointNet to make an hierarchy to learn local features. We will show later that our much simpler approach can still learn local features effectively and achieve the state-of-the-art results.

%------------------------------------------------------------------------
\section{\ournet}
\label{sec:net}

We now proceed to design a convolutional neural network for point cloud feature learning. We draw inspirations from typical 2D convolutional neural networks and build an architecture named \ournet which uses \ourconv in place of traditional 2D convolution (see Figure~\ref{fig:net}).
This architecture can be used for multiple scene understanding tasks. Particularly, the classification and segmentation networks both share the encoder part, and only differ in the part after that. Since \ourconv is permutation invariant for input points, \ournet is able to consume point sets directly.

Our network for point cloud deep learning has three layers. In the classification stage, we pass all the input points through three \ourconv operators. The points are gradually subsampled into less representative points denoted as $N_{0} > N_{1} > N_{2}$ respectively, while the output feature channels increases layer by layer, denoted as $C_{0} < C_{1} < C_{2}$ respectively. In Figure~\ref{fig:net}, $N_{i}$ represented as blue dots with thicker shape that indicates a higher dimension. This design is similar to a typical 2D convolutional neural network: the number of representative points decreases while the number of output channels increases. After three layers of \ourconv, we obtain a matrix of size $N_{2} \times C_{2}$, where $N_{2}$ is the final number of representative points extracted from the input point cloud with each one contains a high dimensional feature vector of size $C_{2}$. This matrix is fed into the \textit{mlp} module size of $(256,128)$ to produce the probability map for object classification. Finally, we obtain a $128 \times k_{cls}$ matrix with $k_{cls}$ indicates the number of classes. The specific parameter settings are discussed in Section~\ref{sub:param_setting}

The segmentation network follows U-net~\cite{ron2015unet}, an encoder-decoder architecture with skip connections. The deconvolution part starts with the set of $N_{2}$ points from the encoder, passing through the \ourconv operators until the point cloud reaches the original resolution. 
%Suppose that in the convolution stage, the point cloud in layer $L_1$ is downsampled and pass to layer $L_2$. Let us now consider deconvolution layer $L_2'$. Note that $L_2'$ and $L_1'$ have the same point clouds as $L_2$ and $L_1$. The features in $L_2'$ is to be transformed to layer $L_1'$. To do so, the points in layer $L_1$ are used as queries, and the neighbor points are searched in the corresponding convolution layer $L_2$. The features in $L_1'$ are assigned accordingly using the same \ourconv procedure.
The deconvolution layers gradually output more points but less feature channels. Skip connections retain  features from
earlier layers and concatenate them to the output features of the deconvolution layers. Such strategy is shown to be highly effective for dense semantic segmentation on images~\cite{ron2015unet}, which we adopts here for point clouds. 
Note that we use \ourconv for both convolution and deconvolution. 
The output $N \times C$ is also fed into an \textit{mlp} to produce the probability map for segmentation, where we obtain a $64 \times k_{seg}$ matrix with $k_{seg}$ indicates the number of segment labels.

%\paragraph{Discussion.}
%Our network architecture is very simple compared to PointNet~\cite{qi2017pointnet} and its extensions~\cite{qi2017pointnet++}. In fact, it is possible to add a PointNet-like architecture to learn global features before feeding their results to our \ourconv. However, empirically we found that this only yields subtle accuracy improvement of $0.5$\% in the object classification task. Therefore, in this work, we opt to keep the network architecture simple with only \ourconv operators.

%-------------------------------------------------------------------------
\section{Experimental Results}
\label{experiment}

In this section, we perform the experiments with three typical point cloud learning tasks: object classification, part segmentation, and semantic segmentation. We evaluate our method under different settings to justify the results. In general, our method achieves the state-of-the-art performance for both accuracy and speed in all the experiments. 

\subsection{Parameter Setting}
\label{sub:param_setting}
\ournet has three encoding layers, each of which contains a \ourconv. The parameters are $N_{i}$, $S_{i}$, and $C_{i}$ that denote the number of representative points, the number of shells, and output channels in each layer respectively. From the first to third layers, $N_{i}$ is set to 512, 128, 32, $S_{i}$ is set to 4, 2, 1, and $C_{i}$ is set to 128, 256, and 512 for $i = 0, 1, 2$ respectively. $C$ is set to 64 at the last convolution for segmentation. 
We define the number of points contained in each shell as shell size, which is set to 16 for classification and 8 for segmentation. So the number of neighbors for each representative point is $S_{i}\times 16$ and $S_{i} \times 8$, which is equal to 64, 32, and 16 for the three layers of classification, and 32, 16, 8 for segmentation, respectively. 
During training, we use a batch size of $32$ for classification and $16$ for segmentation. The optimization is done with an Adam optimizer with initial learning rate set to 0.001. Our network is implemented in TensorFlow~\cite{abadi2016tensorflow} and run on a NVIDIA GTX 1080 GPU for all experiments.

\subsection{Object Classification}
\begin{table}[b]
	\centering
	\footnotesize
	\begin{tabular}{lcccc}
		\toprule
		Method   & Core Operator & input & OA \\
		\midrule
		FPNN \cite{li2016fpnn} & 1D Conv.  &P & 87.5  \\
		Vol. CNN \cite{qi2016volumetric} & 3D Conv.  &V & 89.9     \\
		O-CNN \cite{wang2017cnn}   & Sparse 3D Conv. &O & 90.6     \\
		Pointwise \cite{hua2017point} & Point Conv. &P & 86.1   \\
		PointNet \cite{qi2017pointnet}  & Point MLP  &P & 89.2     \\
		PointNet++ \cite{qi2017pointnet++}   & Multiscale Point MLP &P+N & 90.7    \\
		PointCNN \cite{li2018pointcnn}    & X-Conv &P & 92.2  \\
		\midrule
		\ournet ($ss$=8)     & \ourconv  &P  & 91.0 \\
		\ournet ($ss$=16)    & \ourconv  &P  & \textbf{93.1} \\
		\ournet ($ss$=32)    & \ourconv  &P  & \textbf{93.1} \\
		\ournet ($ss$=64)    & \ourconv  &P  & 92.8 \\
		\bottomrule
	\end{tabular}
	\smallgap
	\caption{Comparisons of classification accuracy (overall accuracy \%) on ModelNet40~\cite{wu20153d} with input type denoted as O (Octrees), V (Voxels), P (Points) and N (Normals). Performance of \ournet is tested with different shell size ($ss$)}
	\label{classification_1}
\end{table}

The classification is tested on ModelNet40 \cite{wu20153d} which is composed of $40$ object classes and has $9,843$ models for training and $2,468$ models for testing. We use the point cloud data of ModelNet40 provided by Qi et al.~\cite{qi2017pointnet} as input, where $1024$ points are roughly uniformly sampled from each mesh. Only the geometric coordinates $(x,y,z)$ of the sampled points are used in the experiment. We follow the train-test split from PointNet~\cite{qi2017pointnet}. The data is augmented by randomly perturbing the point locations. The comparison results are shown in Table~\ref{classification_1}. 

As we can see, our results have achieved the state-of-the-art. While \ournet with shell size ($ss$) of 16 is the default setting for classification, other $ss$ are also tested. When decreasing $ss$ to 8, the receptive fields become smaller and less overlapped and accuracy also decreases slightly but still around $91.0\%$. When $ss$ increases, receptive field is enlarged so that more spatial context information is captured. \ournet achieves $93.1\%$ accuracy with $ss$ is $32$. However, this does not mean the larger the better, since too large receptive field can also wash out the high frequency fine structure of the features~\cite{bartlett1990how}. We can see that when $ss$ is set to $64$, the accuracy drops to $92.8\%$. To balance between speed and accuracy, we set $ss$ to $16$ for object classification.
Figure~\ref{fig:time} provides an accuracy plot under equal-time and equal-epoch setting. As can be seen, our method outperformed all tested methods, being the fastest and most accurate towards convergence.
Compared to PointCNN~\cite{li2018pointcnn}, one of the fastest method in this experiment, we use a much simpler network architecture. To turn the point cloud into a latent canonical representation, their X-Conv operator requires to learn a transformation matrix while our method only requires a statistical computation to aggregate features. This allows our convolution to be more intuitive and easy to implement but able to achieve high performance. We also provide per-class accuracy in the supplementary document.

\subsection{Segmentation}
Segmentation aims to predict the label for each point, which can also be seen as a dense pointwise classification problem. In this subsection, both object part segmentation and semantic scene segmentation are performed. We use ShapeNet dataset \cite{yi2016scalable} for part segmentation, which contains $16,880$ models ($14,006$ models for training and $2,874$ models for testing) in $16$ categories, each annotated with $2$ to $6$ parts and there are $50$ different parts in total. \new{ For semantic segmentation, we use  ScanNet~\cite{dai2017scannet} and S3DIS dataset~\cite{armeni-parsing-cvpr16} for indoor scenes, and Semantic3D~\cite{hackel2017isprs} for outdoor scenes. 
ScanNet consists of 1513 RGB-D reconstructed indoor scenes annotated in 20 categories. S3DIS contains 3D scans from Matterport scanners in 6 indoor areas including 271 rooms with each point is annotated with one of the semantic labels from 13 categories.  Semantic3D is an online large-scale, outdoor LIDAR benchmark dataset comprising more than 4 billion annotated points with 8 classes. We follow PointCNN~\cite{li2018pointcnn} to prepare the datasets.}

\begin{figure}[t]
	\centering
	\includegraphics[width=0.9\linewidth]{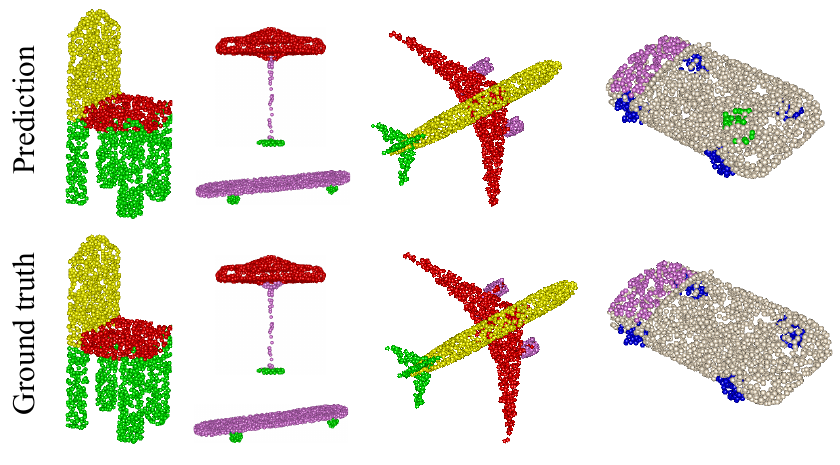}
	\caption{Object part segmentation with the ShapeNet dataset. The example objects are a chair, lamp, skateboard, airplane, and a car. Overall, our method produces accurate predictions.}
	\label{fig:object_part}
\end{figure}

\noindent \textbf{Object Part Segmentation.}
Our results are reported in Table~\ref{tab:segmentation}. Per-class accuracies can be found in the supplementary document. It can be seen that our method outperforms most of the state-of-the-art techniques. Qualitative comparisons between our prediction and the ground truth are shown in Figure~\ref{fig:object_part}. It can be seen that \ournet method can run robustly on many objects. Noted that our method only trains $20$ hours to achieve such accuracy.
\begin{table}[t]
	\begin{center}
	\footnotesize
	\begin{tabular}{l c|c|c|c}
		\toprule
		Method   & ShapeNet & ScanNet & S3DIS  & Semantic3D \\
		         & mpIoU    & OA      & mIoU   & mIoU \\
		\midrule
		SyncCNN \cite{yi2017syncspeccnn}     & 82.0   & - & - & -\\
		SpiderCNN \cite{xu2018spidercnn}     & 81.7   & - & - & -\\
		SplatNet \cite{su18splatnet}     & 83.7   & - & - & -\\
		SO-Net \cite{li2018sonet}     & 81.0   & -   & - & -\\
		SGPN \cite{wang2018sgpn}     & 82.8   & - & 50.4 & -\\
		PCNN \cite{atzmon2018point}     & 81.8   & -  & - & -\\
		KCNet \cite{shen2018mining}     & 82.2   & -    & - & -\\
		KdNet \cite{klokov2017escape}     & 77.4   & - & - & -\\
		3DmFV-Net \cite{shabat20183dmfv} & 81.0  & - & - & -\\
		DGCNN \cite{wang2018edgeconv}     & 82.3   & - & 56.1 & -\\
		RSNet \cite{huang2018recurrent}     & 81.4   & - & 56.5 & -\\
		PointNet \cite{qi2017pointnet}     & 80.4   & 73.9 & 47.6 & -\\
		PointNet++ \cite{qi2017pointnet++} & 81.9   & 84.5     & -  & -\\
		PointCNN \cite{li2018pointcnn}     & \textbf{84.6}   & 85.1 & 65.4   & -\\
		TMLC-MSR~\cite{hackel2016fast}     & -   & - & -  & 54.2 \\
		DeePr3SS~\cite{lawin2017deep}      & -   & - & -  & 58.5 \\
		SnapNet~\cite{boulch2017unstructured}  & -   & - & -  & 59.1\\
		SegCloud~\cite{tchapmi2017segcloud}    & -   & - & -  & 61.3\\
		SPG~\cite{landrieu-superpoint-cvpr18}  & -   & - & 62.1  & \textbf{73.2}\\
		\midrule
		Ours     &  82.8 &  \textbf{85.2}  & \textbf{66.8} & 69.4\\
		\bottomrule
	\end{tabular}
	\smallgap
	\caption{Comparisons of segmentation tasks. Object part segmentation is performed on ShapeNet dataset~\cite{chang2015shapenet}, and semantic segmentation is performed on ScanNet~\cite{dai2017scannet}, S3DIS dataset~\cite{armeni-parsing-cvpr16}, and Semantic3D~\cite{hackel2017isprs} respectively.}
	\label{tab:segmentation}
	\end{center}
\end{table}

\begin{figure*}[t]
	\centering
	\includegraphics[width=1.0\linewidth]{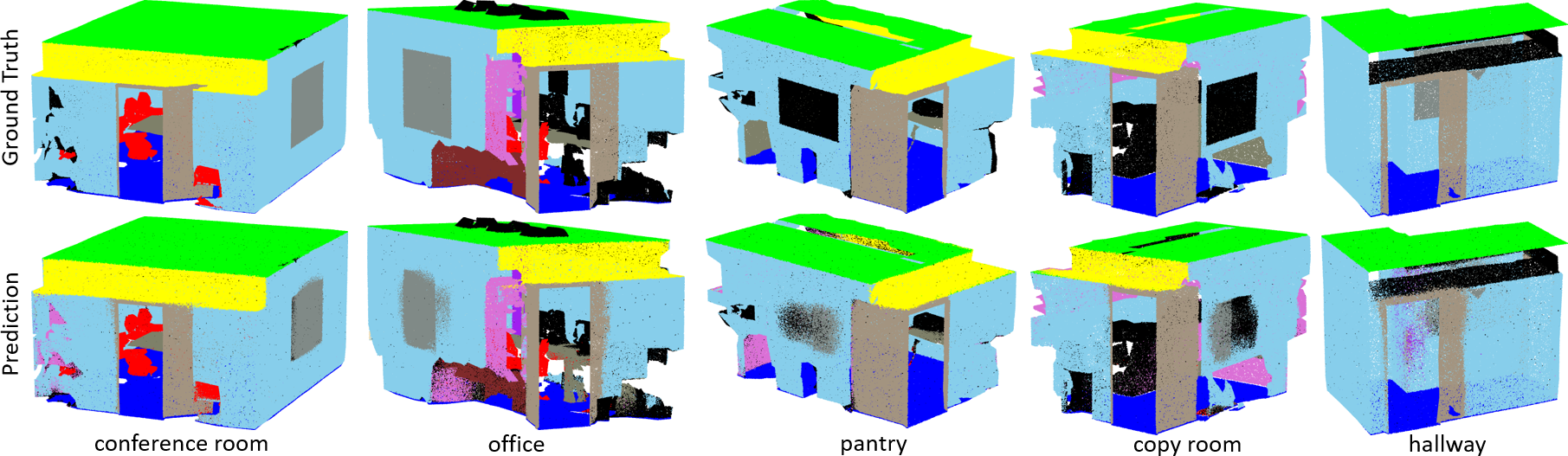}
	\caption{Semantic segmentation for indoor scenes in the S3DIS dataset~\cite{armeni-parsing-cvpr16}.}
	\label{fig:semantic_seg}
\end{figure*}

\noindent \textbf{Indoor Semantic Scene Segmentation.} 
The mIoU accuracies of indoor benchmarks ScanNet~\cite{dai2017scannet} and S3DIS~\cite{armeni-parsing-cvpr16}  are shown in Table~\ref{tab:segmentation}. \ournet ranks 1st on ScanNet and ranks 1st on S3DIS. For the latter, we also list the per-class scores (mIoU) in the supplementary document. The qualitative results are presented in Figure~\ref{fig:semantic_seg}. We can see some misclassification are between wall, caseboard, and window as these categories are quite similar in pure geometries, and need other features like color or normal vectors to improve. 

\begin{figure*}
	\centering
	\includegraphics[width=.48\linewidth,height=85pt]{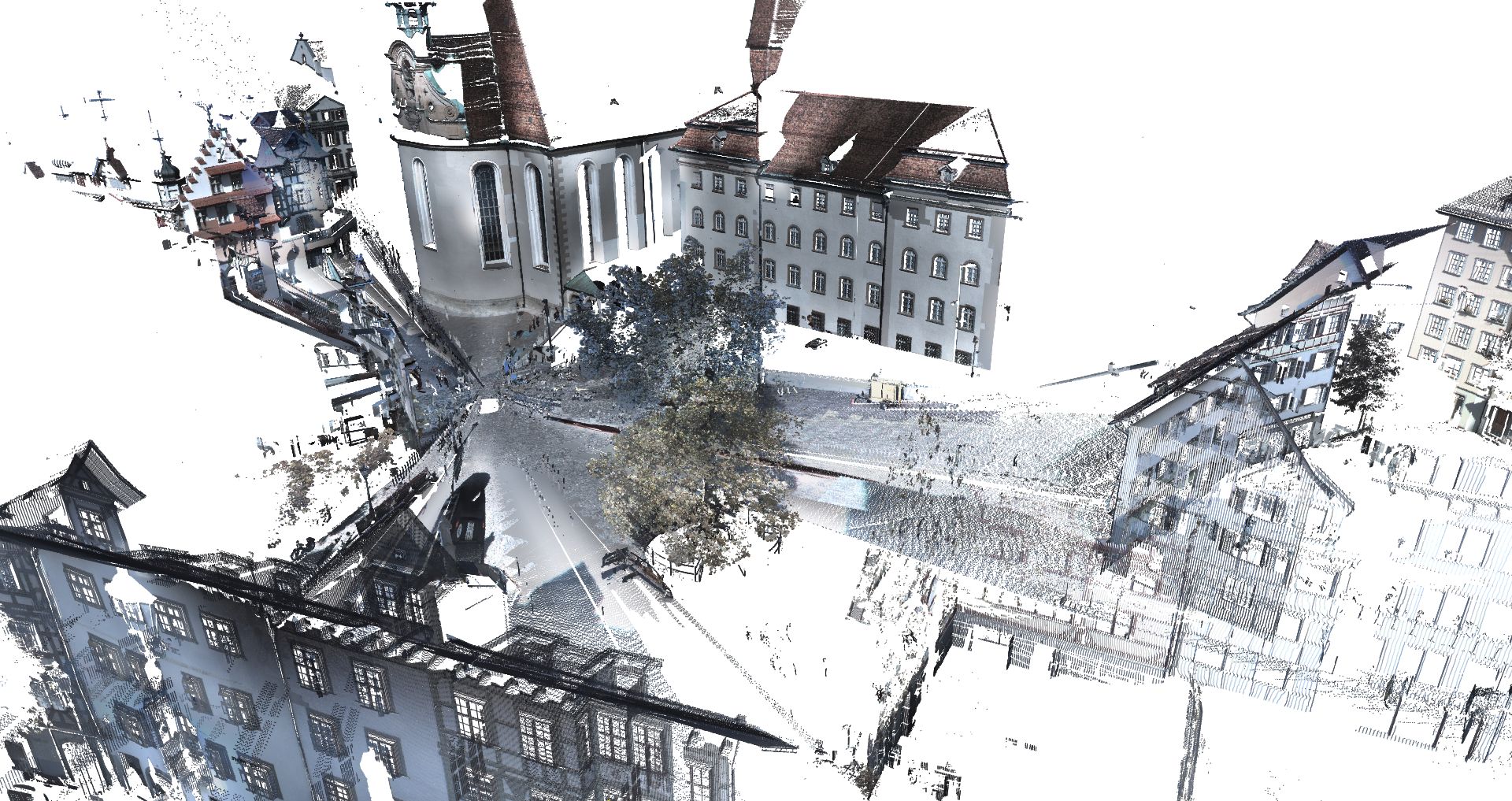}%
	\hspace{.15cm}%
	\includegraphics[width=.48\linewidth,height=85pt]{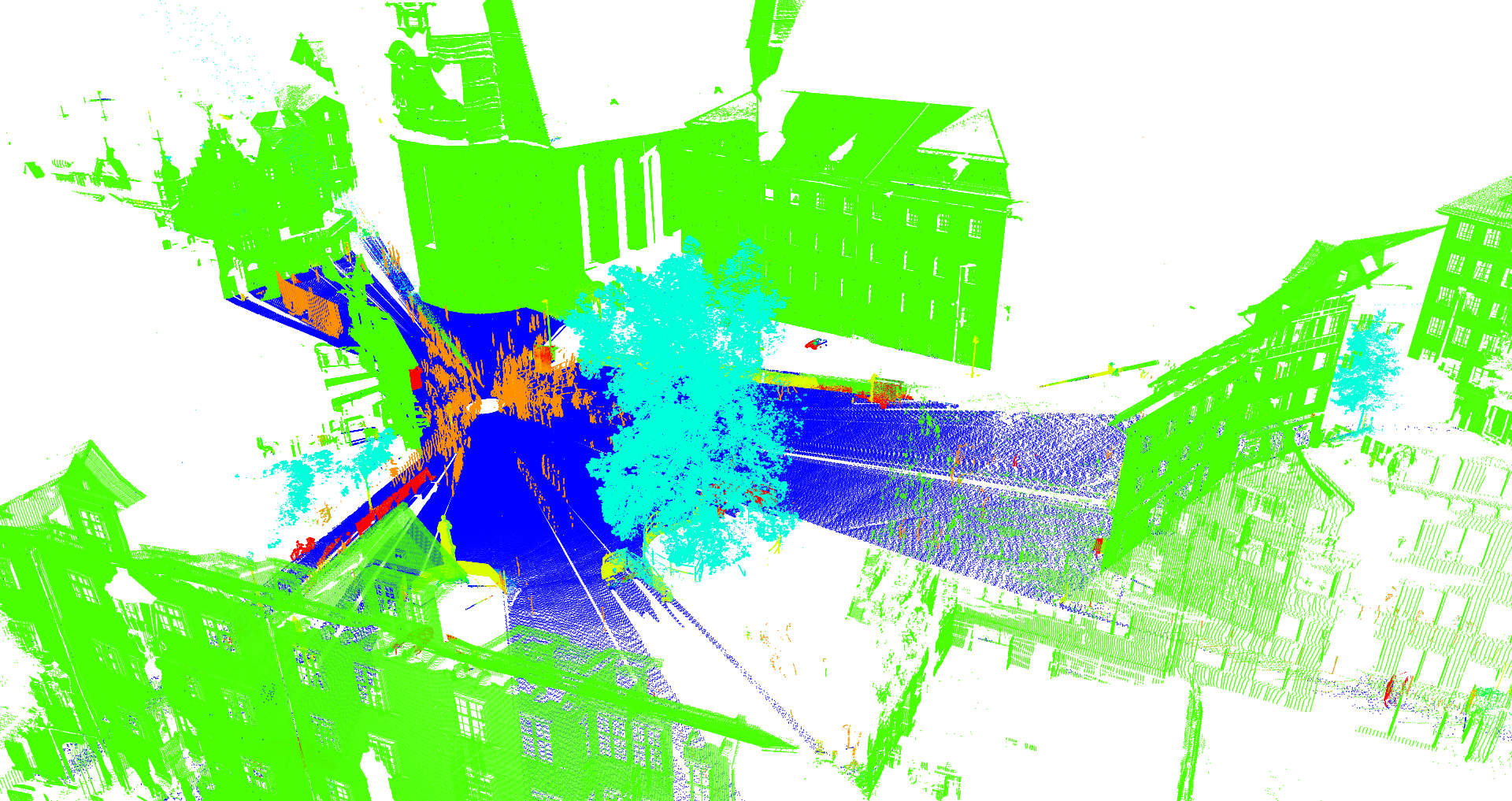}%
	\vspace{.05cm}
	\includegraphics[width=.48\linewidth,height=75pt]{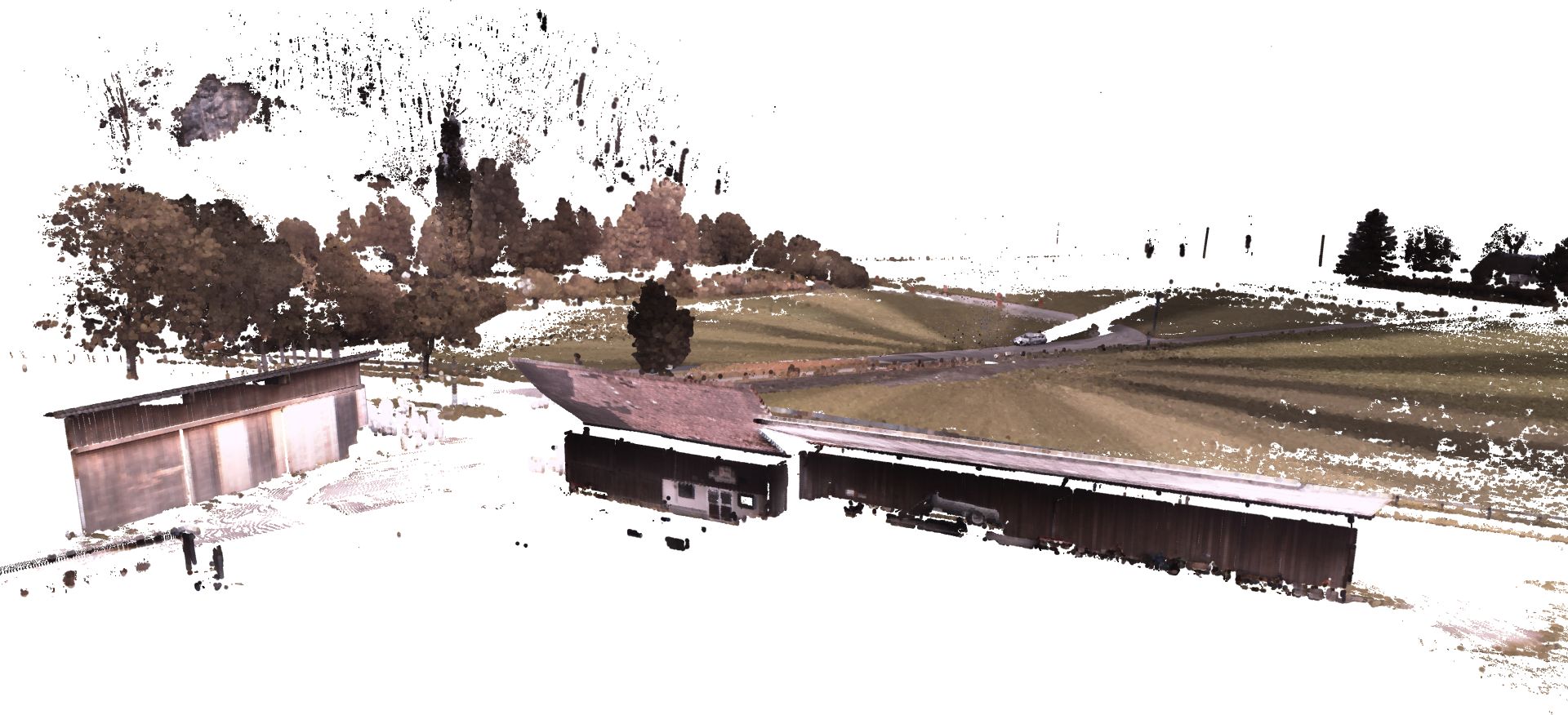}%
	\hspace{.15cm}%
	\includegraphics[width=.48\linewidth,height=75pt]{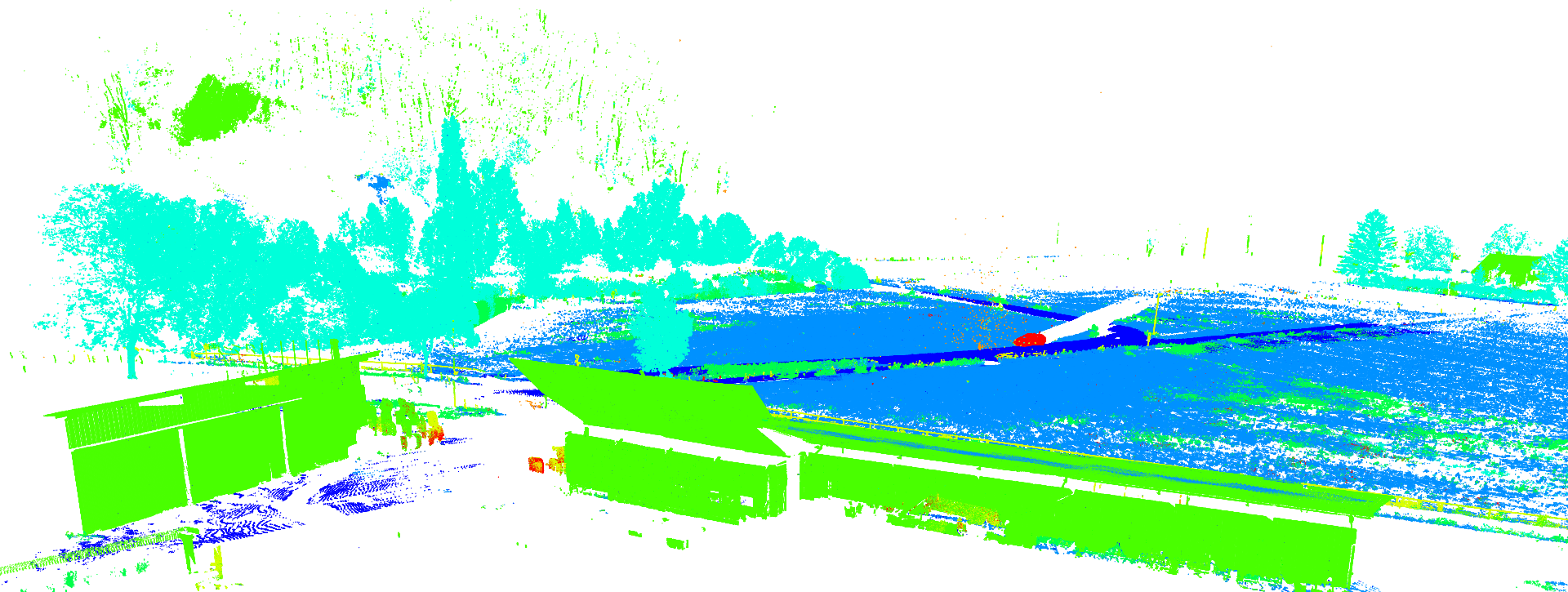}%
	\caption{Semantic segmentation for outdoor scenes in the Semantic3D dataset~\cite{hackel2017isprs}. Left: colored point clouds (for visualization only). Right: our segmentation. Note that the ground truth of the test set is not publicly available.}
	\label{fig:semantic3d_seg}
\end{figure*}

\noindent \textbf{Outdoor Semantic Scene Segmentation.} 
\new{The Semantic3D~\cite{hackel2017isprs} is more challenging as it is a real-world dataset of strongly varying point density. For a fair comparison, we excluded results without a publication. \ournet performs well on this dataset with accuracy ranked 2nd (Table~\ref{tab:segmentation}). Per-class accuracy can be found in the supplementary document. The qualitative results are presented in Figure~\ref{fig:semantic3d_seg}. Note that our method only takes the 3D coordinates as input while previous methods such as~\cite{landrieu-superpoint-cvpr18} also used color or postprocess with CRFs.}
%\begin{figure*}
%  \centering
%  \begin{tabular}{@{}c@{}}
%    \includegraphics[width=.48\linewidth,height=100pt]{example-image-a} \\[\abovecaptionskip]
%    \small (a) An image
%  \end{tabular}
%
%%  \vspace{\floatsep}
%
%  \begin{tabular}{@{}c@{}}
%    \includegraphics[width=.48\linewidth,height=100pt]{example-image-b} \\[\abovecaptionskip]
%    \small (b) Another image
%  \end{tabular}
%
%  \caption{This is a figure caption}\label{fig:myfig}
%\end{figure*}

%\begin{figure*}[t]
%	\centering
%	\includegraphics[width=1.0\linewidth, height=0.15\textheight]{figures/scannet/scannet_0.png}
%	\caption{Semantic segmentation with Semantic3D dataset.}
%	\label{fig:semantic3d}
%\end{figure*}

\subsection{Network Efficiency} 
\label{sub:efficiency}
We measure network complexity by the number of trainable parameters, floating point operations (FLOPs), and running time to analyze the network efficiency. With batch size 16, point cloud size 1024 from the ModelNet40 dataset, the statistics are reported in Table~\ref{tab:parameters}. For all three metrics, \ournet is better than existing methods. \new{While being much less complex in time and space, \ournet can still converge to the state-of-the-art accuracy very efficiently as shown in the plot in Figure~\ref{fig:time}.}

\new{
The improvement in speed and memory of our work comes from the effective use of \textit{mlp} and 1D convolution in our network. 
Particularly, on top of the proposed systematic approach based on concentric shells for point grouping, which naturally handles multi-scale features, we only need a single \textit{mlp} to learn point features in a shell, and a 1D convolution to relate features among the shells (Figure~\ref{fig:conv}). This simplicity greatly reduces the number of trainable parameters and computation.
}
%It can be explained by (1) our larger receptive field allows network to see more spatial distribution from the input without increasing the number of network layers, and (2) the convolution order is determined naturally through \ourconv structure without the need of extra learning~\cite{li2018pointcnn}, transformation~\cite{qi2017pointnet, qi2017pointnet++} or special representations~\cite{su2015multi,qi2016volumetric}. 
%But this does not mean the larger the better as we see from the results of Table~\ref{classification_1}. 

\begin{figure}[t]
	\includegraphics[width=0.95\linewidth]{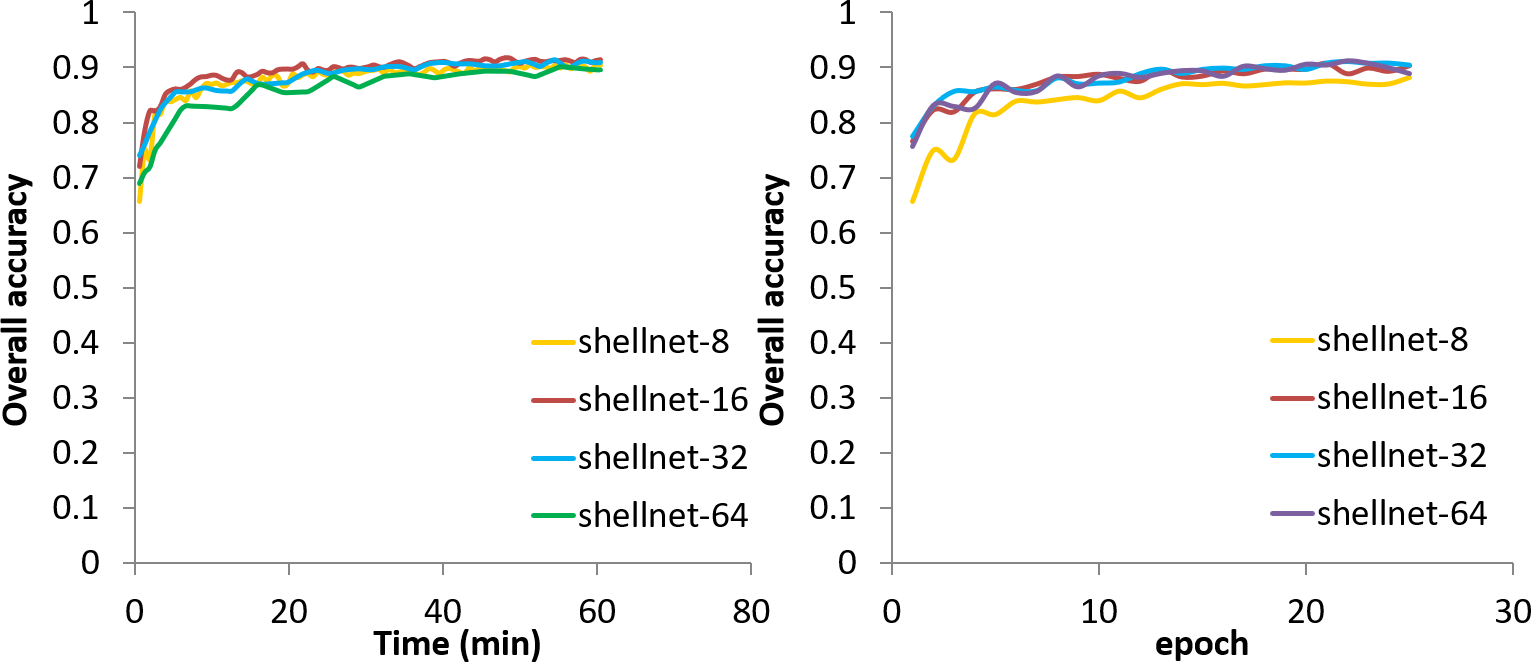}
	\caption{The accuracy of point cloud classification versus time and epochs with different shell sizes.} 
	\label{fig:acc_time_epoch_shellnet}
\end{figure}

In \ournet, the receptive field is directly controlled by the shell size. Thus, we can further analyse the performance of \ournet with different shell sizes (Figure~\ref{fig:acc_time_epoch_shellnet}). In equal-time comparison, ShellNet with shell size 16 performs best, achieving high accuracy in very short time. When shell size is 64, it performs slightly worse. In equal-epoch setting, using shell size 8 is not as good as the others because of smaller receptive fields. Having a receptive field that balances between size and speed yields the best convergence. 
\begin{table}[t]
  \begin{center}
  \footnotesize
    \begin{tabular}{lc|c|c} % <-- Alignments: 1st column left, 2nd middle and 3rd right, with vertical lines in between
      \toprule
      Methods &Params & FLOPs & Time\\
       &  & (Train/Infer) & (Train/Infer)\\
      \midrule
      PointNet \cite{qi2017pointnet} & 3.5M & 44.0B / 14.7B & 0.068s / 0.015s \\
    
      PointNet++ \cite{qi2017pointnet++}& 12.4M & 67.9B /26.9B & 0.091s / 0.027s \\
    
      3DmFV \cite{shabat20183dmfv}& 45.77M & 48.6B /16.9B & 0.101s / 0.039s \\
     
      DGCNN \cite{wang2018edgeconv} & 1.84M & 131.4B /44.3B & 0.171s / 0.064s \\

      PointCNN \cite{li2018pointcnn} & 0.6M & 93.0B /25.3B & 0.031s / 0.012s \\
      \midrule
      \ournet & \textbf{0.48M} & \textbf{15.8B} /\textbf{2.8B} & 0.066s / 0.023s \\
      ~~~~with small RF & \textbf{0.48M} & \textbf{9.51B} /\textbf{1.5B} & \textbf{0.025s} / \textbf{0.011s} \\
      \bottomrule
    \end{tabular}
  \end{center}
  \caption{Trainable parameters, FLOPs and running time comparisons. Compared to previous methods, ShellNet is lightweight and fast while being accurate. Reducing the receptive field (small RF)
  %, e.g. be consistent with that of PointCNN~\cite{li2018pointcnn}) 
  by setting a smaller shell size can make the computation even faster as neighbor query becomes cheaper.}
  \label{tab:parameters}
\end{table}

\subsection{Neighboring Point Sampling}
\new{
Let the network in Figure~\ref{fig:net} as the baseline (Setting A), we conduct a series of experiments to verify the effectiveness of our network architecture and justify how neighbor points can be sampled. 
}

\new{
Here we compare four settings. 
Particularly, Setting A is the default configuration of our classification experiment, in which random sampling is used to obtain neighbor points and each concentric shell contains a fixed number of points. Setting B, C, D are obtained by varying the neighbor sampling strategies. In Setting B, we change neighbor sampling to farthest point sampling. 
In Setting C, we divide a local region into equidistant shells, leading to shells that contain a dynamic number of points. In Setting D, we search nearest neighbors in the feature space instead of the 3D coordinate space.
The results are shown in Table~\ref{tab:ablation}.
It shows that accuracies are similar across variants, and Setting A is the most efficient for the classification task. For segmentation, we also conduct the same experiment and found that Setting B works best in this case. The reason is that farthest point sampling in Setting B results in more uniform point distribution that can cover more geometry details, leading to more accurate segmentation. 
}
\begin{table}[t]
	\footnotesize
	\centering
	\begin{tabular}{l|c|c|c|c}
	\toprule
		& (A) & (B) & (C) & (D) \\
	 1. Sampling & Random & \textbf{Farthest} & Random & Random \\
	2. Shell size & Fixed & Fixed & \textbf{Dynamic} & Fixed  \\
	3. KNN type & $xyz$ & $xyz$ & $xyz$ & \textbf{Features}  \\
	\midrule
	Accuracy (\%) & 93.1 & 93.1 & 92.7 & 92.4  \\
	Train Time & 0.066s & 0.078s & 0.118s & 0.081s  \\
	Infer. Time & 0.023s & 0.024s & 0.033s & 0.029s  \\
	\bottomrule
	\end{tabular}
	\smallgap
	\caption{Experiments with neighbor point sampling. Setting (A) is the default strategy. Setting (B), (C), (D) are modified from (A) based on point sampling type, shell size, and neighbor query features. As can be seen, setting (B) -- furthest point sampling, (C) -- equidistant shells, (D) -- latent features for neighborhood construction, produces similar accuracy but training and inference time becomes slower.}
	\label{tab:ablation}
\end{table}

\begin{figure}[t]
	\centering
	\includegraphics[width=\linewidth]{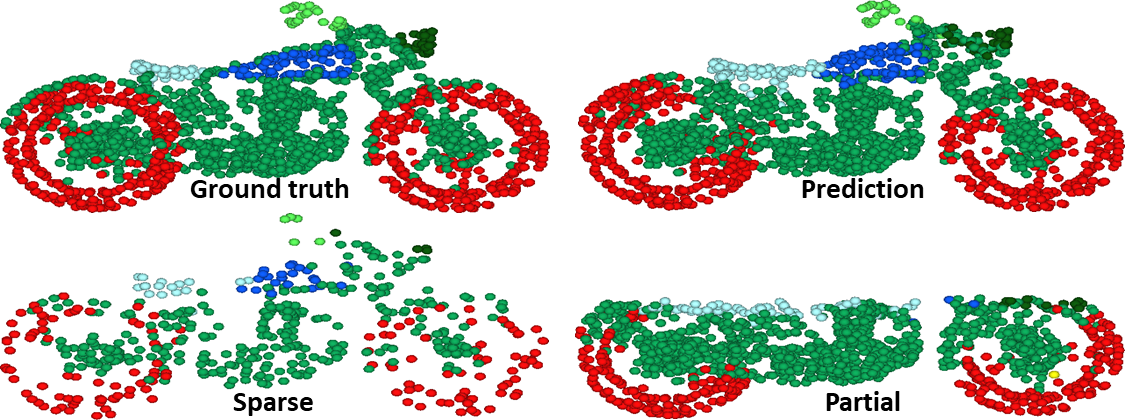}
	\caption{Part segmentation on sparse and partial point clouds. For partial data, points on the boundaries appear to be less accurate.}
	\label{fig:motorbike}
\end{figure}

%\cCH{Do we use KNN or radius here?}
\new{
%\paragraph{Limitation.}
Our method is without limitation. Particularly, we found that while our method can work with sparse and partial data, more investigations into its robustness is required. Here we provide an example of object part segmentation to demonstrate the robustness of \ourconv in Figure~\ref{fig:motorbike}. 
The mpIoU accuracies for the original, sparse and partial segmentation are 82.4\%, 80.2\%, 72.6\% respectively. 
For partial data, the boundary points are less accurate.
}

%-------------------------------------------------------------------------
\section{Conclusion}
\label{conclusion}
We introduced a novel approach for deep learning with 3D point clouds based on concentric spherical shells constructed from local point sets. We designed a new convolution operator named, \ourconv, which supports convolution of a point set efficiently based on shells and their statistics. This structure not only solves the convolution order problem naturally but also allows larger and more overlapped receptive field without increasing the number of network layers. Based on \ourconv, we build simple yet effective neural network that achieves the state-of-the-art results on object classification and segmentation tasks with pure point cloud inputs.

Together with recent advances in deep learning with point cloud data, our work leads to several potential future research. With the fast capability of local feature learning, it would be interesting to see how object detection and semantic instance segmentation can benefit from our work. It is also interesting to extend this work for learning with meshes. 
Finally, it would be of great interest to apply our approach to build autoencoders for point cloud generation.

\noindent
\textbf{Acknowledgements.} The authors acknowledge support from the SUTD Digital Manufacturing and Design Centre  funded by the Singapore National Research Foundation, and an internal grant from HKUST (R9429).

{\small
\bibliographystyle{ieee_fullname}
\bibliography{egbib}
}

\end{document}